\pdfoutput=1

\documentclass[11pt]{article}

\usepackage{acl2023}

\usepackage{times}
\usepackage{latexsym}

\usepackage[T1]{fontenc}

\usepackage[utf8]{inputenc}

\usepackage{microtype}

\title{Code-Switched Text Synthesis in Unseen Language Pairs}
\author{
    I-Hung Hsu\textsuperscript{\rm 1}\thanks{\; Work was done when the author interned at Amazon.} \ \ \ 
    Avik Ray\textsuperscript{\rm 2} \ \ \ 
    Shubham Grag\textsuperscript{\rm 2} \ \ \
    Nanyun Peng\textsuperscript{\rm 2,3} \ \ \
    Jing Huang\textsuperscript{\rm 2} \\
    \textsuperscript{\rm 1} Information Science Institute, University of Southern California \\ \textsuperscript{\rm 2} Amazon Alexa AI \ \ \ \  
    \textsuperscript{\rm 3} University of California, Los Angeles \\
    \texttt{ihunghsu@usc.edu} \ \ \ \texttt{\{avikray, gargshu\}@amazon.com} \\
    \texttt{violetpeng@cs.ucla.edu} \ \ \ \texttt{jinghuang.zhu@gmail.com}
}
\usepackage{graphicx}
\usepackage{multirow}
\usepackage{makecell}
\usepackage{booktabs}
\usepackage{enumitem}
\usepackage{amssymb}
\usepackage{xcolor}
\usepackage{color}
\usepackage{colortbl}
\usepackage{multirow}
\definecolor{dark-green}{rgb}{0.31, 0.47, 0.26}
\definecolor{dark-red}{rgb}{0.81, 0.09, 0.13}

\usepackage{soul}
\usepackage{amsmath}

\setuldepth{Berlin}
\usepackage{tabularx}
\newcolumntype{x}[1]{>{\arraybackslash\hspace{0pt}}m{#1}}

\usepackage{pifont}

\newcommand{\model}{\textsc{Gloss}}

\newcommand{\mbf}[1]{\mathbf{#1}}

\usepackage{todonotes}

\usepackage{cleveref}
\crefformat{section}{\S#2#1#3}
\crefformat{subsection}{\S#2#1#3}
\crefformat{subsubsection}{\S#2#1#3}
\crefrangeformat{section}{\S\S#3#1#4 to~#5#2#6}
\crefmultiformat{section}{\S\S#2#1#3}{ and~#2#1#3}{, #2#1#3}{ and~#2#1#3}
\crefmultiformat{subsection}{\S\S#2#1#3}{ and~#2#1#3}{, #2#1#3}{ and~#2#1#3}
\Crefformat{figure}{#2Fig.~#1#3}
\Crefmultiformat{figure}{Figs.~#2#1#3}{ and~#2#1#3}{, #2#1#3}{ and~#2#1#3}
\Crefformat{table}{#2Tab.~#1#3}
\Crefmultiformat{table}{Tabs.~#2#1#3}{ and~#2#1#3}{, #2#1#3}{ and~#2#1#3}
\Crefformat{appendix}{Appendix~\S#2#1#3}
\crefmultiformat{appendix}{Appendix~\S#2#1#3}{ and~#2#1#3}{, #2#1#3}{ and~#2#1#3}
\crefformat{algorithm}{Alg.~#2#1#3}
\Crefformat{equation}{Eq.~#2#1#3}

\begin{document}
\maketitle

\begin{abstract}
Existing efforts on text synthesis for code-switching mostly require training on code-switched texts in the target language pairs, limiting the deployment of the models to cases lacking code-switched data.
In this work, we study the problem of synthesizing code-switched texts for language pairs absent from the training data.
We introduce \model{}, a model built on top of a pre-trained multilingual machine translation model (PMMTM) with an additional code-switching module. 
This module, either an adapter or extra prefixes, learns code-switching patterns from code-switched data during training, while the primary component of \model{}, i.e., the PMMTM, is frozen.
The design of only adjusting the code-switching module prevents our model from overfitting to the constrained training data for code-switching.
Hence, \model{} exhibits the ability to generalize and synthesize code-switched texts across a broader spectrum of language pairs. 
Additionally, we develop a self-training algorithm on target language pairs further to enhance the reliability of \model{}. 
Automatic evaluations on four language pairs show that \model{} achieves at least 55\% relative BLEU and METEOR scores improvements compared to strong baselines. Human evaluations on two language pairs further validate the success of \model{}. 

\end{abstract}
\section{Introduction}
\label{sec:intro}
Code-switching, the linguistic phenomenon of using more than one language within a single utterance or conversation,\footnote{In this paper, we mainly focus on the sentence-level code-switching involving only two languages.} is a common expression of multilingualism in informal text and speech~\cite{auer2008handbook, gumperz1982discourse}.
To accommodate the needs of multicultural and multilingual societies and individuals, there is a growing interest in investigating models dedicated to code-switching within the realm of conversational AI~\cite{Massive2022Jack, GLUECos2020Simran, DecadesSurvey2022Winata, DBLP:journals/corr/abs-1904-00784}.
However, a notable obstacle in code-switching modeling is the scarcity of large-scale code-switched text datasets for different applications in diverse language pairs~\cite{gupta2020semi, tarunesh2021machine}.
This necessitates generative models capable of synthesizing code-switched texts, facilitating subsequent studies for code-switching.

\begin{figure}[t!]
    \centering
    \includegraphics[trim=0cm 0cm 0cm 0cm, clip, width=0.99\columnwidth]{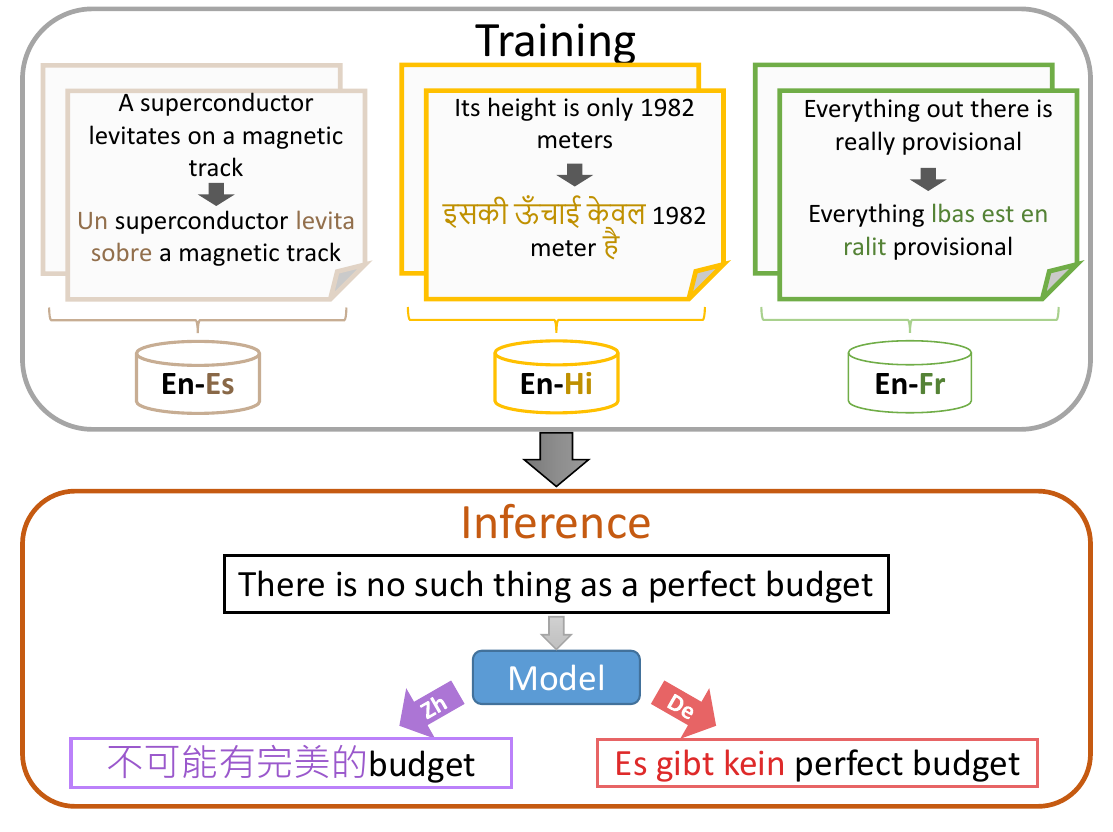}
    \vspace{-0.4em}
    \caption{An illustration of our problem setting. Given a sentence in arbitrary languages (English \textit{(En)} in this figure) and a designated language (Chinese \textit{(Zh)} and German \textit{(De)},  in the figure), the model needs to synthesize a corresponding code-switched sentence that mixes \textit{the original language} and \textit{the designated language}. Additionally, we allow the designated language selections to differ from examples seen during training. \looseness=-1}
    \label{fig:formulation}
\end{figure}

Most prior work on text synthesis for code-switching assumes the availability of training data for all language pairs being tested.
Early trials concentrate on individual language pair~\cite{samanta2019deep, chang2018code, tarunesh2021machine}.
For example, \citet{DBLP:journals/corr/BhatCB16} develop a code-switched text synthesizer for Hindi-English based on linguistic rules~\cite{poplack1980sometimes, belazi1994code, myers1997duelling}, while \citet{lee2019linguistically, winata2019code, DBLP:conf/emnlp/GargPJ18} explore neural generative models for Chinese-English code-switched text synthesis. 
More recently, \citet{gupta2020semi} presents pioneering efforts in developing a generic method for producing high-quality and fluent code-switched sentences across diverse language pairs. This is achieved through the collection of code-switched texts in multiple languages. 

However, the requirement of training on code-switched texts for target language pairs hinders the scalability of existing models to cover a broader range of language pairs. 
Many real-world code-switching scenarios, such as Swahili-English in Tanzania~\cite{kanijo2018code}, Shona-English in Zimbabwe~\cite{mashiri2002shona} suffer from limited or non-existent curated datasets. 
Recognizing this resource limitation, in this work, our study focuses on synthesizing code-switched text in multiple language pairs, including those language pairs that are \textit{unseen} during training (\textit{zero-shot transfer} setting~\cite{Huang2021robust-xlt, acl2022xgear}). 
In this setting, models must learn code-switched patterns from limited code-switched training data in some language pairs and generalize to other language pairs, as shown in \Cref{fig:formulation}.
The setting enables a more flexible process of code-switched text synthesis by using existing resources to assist resource-limited language pairs.
Yet, it also introduces new challenges: (1) models must possess the ability to generate tokens in multiple languages; (2) models need to acquire a transferable ability for code-switching such that they can generate code-switched text in unseen language pairs.

To overcome the challenges, we propose \model{}, a \textbf{G}enera\textbf{L}ized c\textbf{O}de-\textbf{S}witched text \textbf{S}ynthesizer that introduces an additional code-switching module to a pre-trained multilingual machine translation model (PMMTM).
The code-switching module, implemented either through an adapter~\cite{adapter} or extra prefixes~\cite{prefixtuning}, offers a parameter-efficient approach to transfer learning from machine translation to code-switched text synthesis.
Inheriting the ability of PMMTM, \model{} can generate text across multiple languages.
The incorporation of an additional code-switching module, instead of directly fine-tuning the PMMTM, serves as an effective method to prevent models from overfitting to the specific training code-switched language pairs.

Furthermore, we develop a self-training algorithm on the target language pairs to improve \model{} further. 
Specifically, our preliminary study shows that although \model{} can successfully generate reasonable code-switched sentences, when performing zero-shot transfer to unseen language pairs, it may still generate non-code-switched sentences (around 11\% to 13\% of cases). 
The proposed self-training framework aims to introduce weakly-supervised signals to help \model{} more stably generate target domain cases when the target language pair is known.\footnote{For example, we know the target scenario is to synthesize Bengali-English code-switched text, despite no Bengali-English code-switched training data being available.} 
To achieve this, we iteratively fine-tune \model{} on a \textit{filtered} dataset that is generated by \model{} itself in the target domain case. The filter incorporates a language identification model to remove low-quality instances.\footnote{The language identification model is trained without using any code-switched data. More details are given in \Cref{subsec:self-training}.}
Being fine-tuned on filtered data, \model{} learns to generate texts that satisfy the filtering rules and become more stable.

Our contribution is three-fold. First, we present \model{}, a code-switched text synthesizer that can generate code-switched sentences across multiple language pairs, even those not in the training data. To the best of our knowledge, we are the first to study this setting. 
Second, we introduce a self-training framework to further improve \model{} under the setting where the target language pair is known. 
Third, extensive experiments, including automatic evaluations on four languages and human evaluations on two languages, showcase \model{}'s strong performance. 
\model{} achieves at least 55\% relative BLEU and METEOR score improvements compared to strong baselines.

\section{Problem Formulation}
\begin{figure*}[t!]
    \centering
    \includegraphics[trim=0cm 0cm 0cm 0cm, clip, width=0.95\textwidth]{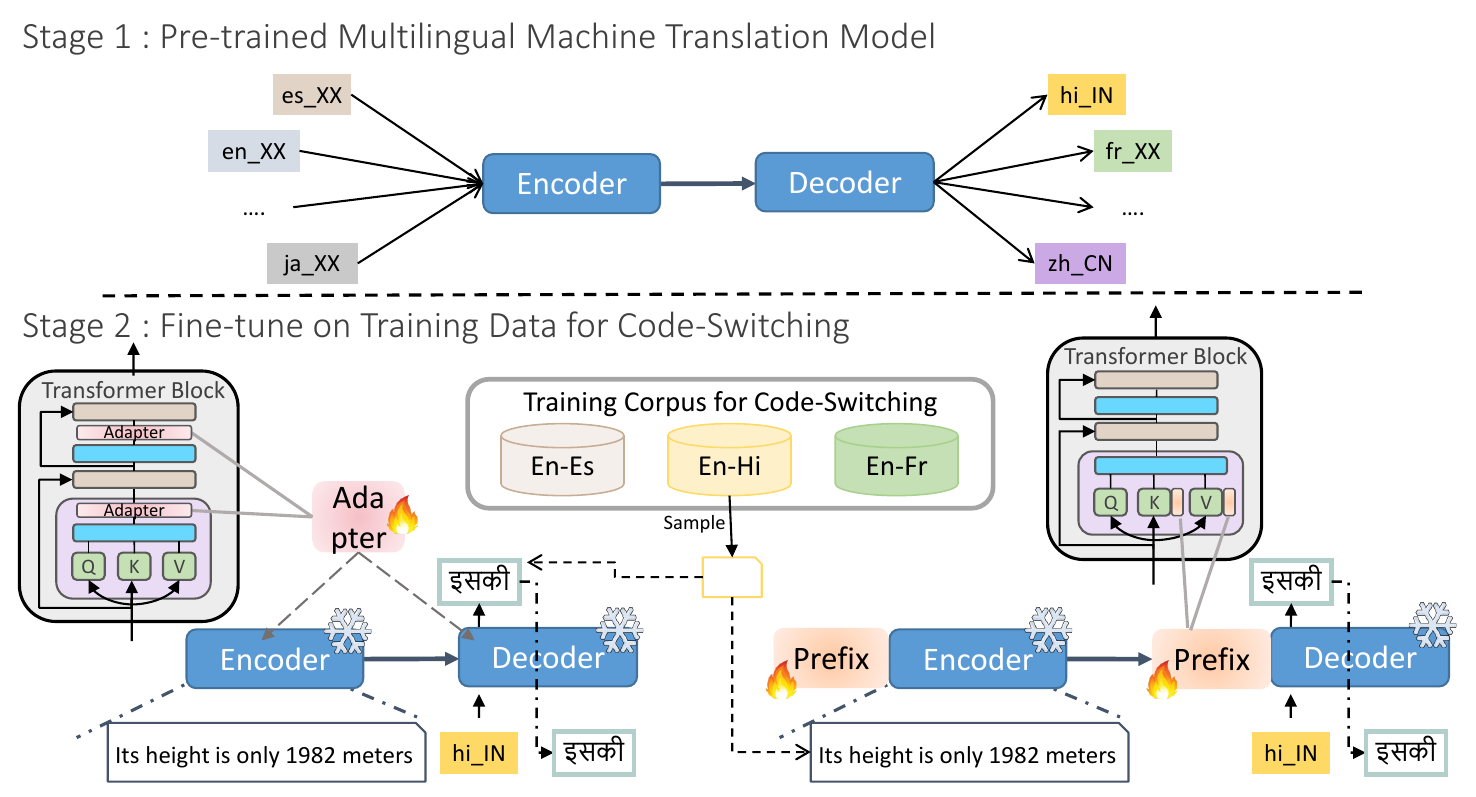}
    \caption{An overview of our \model{} model. \model{} is built on top of a pre-trained multilingual machine translation model, which is trained using machine translation data in many different language pairs. After the pre-trained multilingual machine translation model is prepared, we augment an adapter or extra prefixes to the model. The adapter or prefixes are trained using code-switched data while the pre-trained multilingual machine translation model's parameter will be frozen during the fine-tuning.}
    \label{fig:overview}
\end{figure*} 
Our goal is to synthesize code-switched (CS) texts for language pairs where their CS examples are \textit{never} provided during training. 

Given a monolingual input sentence $\mbf x^e$ in language $l^e$ and an assigned language $l^m$ ($l^m \neq l^e$), we aim to generate a sentence $\mbf x^{m,e}$ that mixes $l^m$ and $l^e$, while preserving the semantic meaning of $\mbf x^e$.\footnote{Following the matrix language frame theory~\cite{myers1997duelling, DBLP:conf/coling/Joshi82}, $l^m$ is called the \ul{m}atrix language and $l^e$ is the \ul{e}mbedded language.} 
We consider the setting where the assigned language $l^m$ in the testing time is different from those in the training time. More formally, as illustrated in Figure~\ref{fig:formulation}, the training set consists of $N$ language pairs $(l^{e_n}, l^{m_n}) (n \in \{1, 2, ..., N\}$), while the testing set includes target language pairs where $l^{m_t} \notin \{l^{m_1}, ..., l^{m_N}\}, \forall t$.
This scenario reflects real-world situations where code-switched data is more readily available for certain language pairs, such as Spanish-English and Hindi-English, while it is less accessible for others, such as Bengali-English and Swahili-English.

\section{Method}
\label{sec:method}
We introduce \model, a \textbf{G}enera\textbf{L}ized c\textbf{O}de-\textbf{S}witched text \textbf{S}ynthesizer that tackles the two specific challenges raised by our problem setting: (1) the model needs to generate texts across many languages, some are not even in the CS training data; (2) the model needs to learn transferable CS ability such that they generate reasonable CS sentences in unseen language pairs. 
\Cref{fig:overview} provides an overview.

To address the first challenge, we begin by obtaining a Pre-trained Multilingual Machine Translation Model (PMMTM) using multilingual machine translation data, which covers all languages that would be used for final CS text synthesis (\Cref{subsec:PMMT}).\footnote{Here, we assume that machine translation data is more available, which is often the case in practice.} 
The remaining challenge is how to make PMMTM a code-switched text synthesizer with only limited language coverage of training data.

We propose to augment an additional code-switching module onto PMMTM, thereby creating \model{} (\Cref{subsec:model}).
This additional code-switching module is trained on our limited CS data while keeping PMMTM parameters fixed. Instead of fine-tuning the entire PMMTM, this modularized design improves systematic generalization~\cite{DBLP:conf/iclr/BahdanauMNNVC19, DBLP:journals/corr/abs-2202-10745}, where PMMTM focuses on generating translated sentences and the code-switching module concentrates on \textit{``mixing''} languages.
This approach allows \model{} to be more adaptable and less prone to overfitting during the fine-tuning process on CS data.

Finally, we present a self-training framework that enables \model{} to more stably generate CS texts in target language pairs (\Cref{subsec:self-training}).

\begin{figure*}[t!]
    \centering
    \includegraphics[trim=0cm 0cm 0cm 0cm, clip, width=0.95\textwidth]{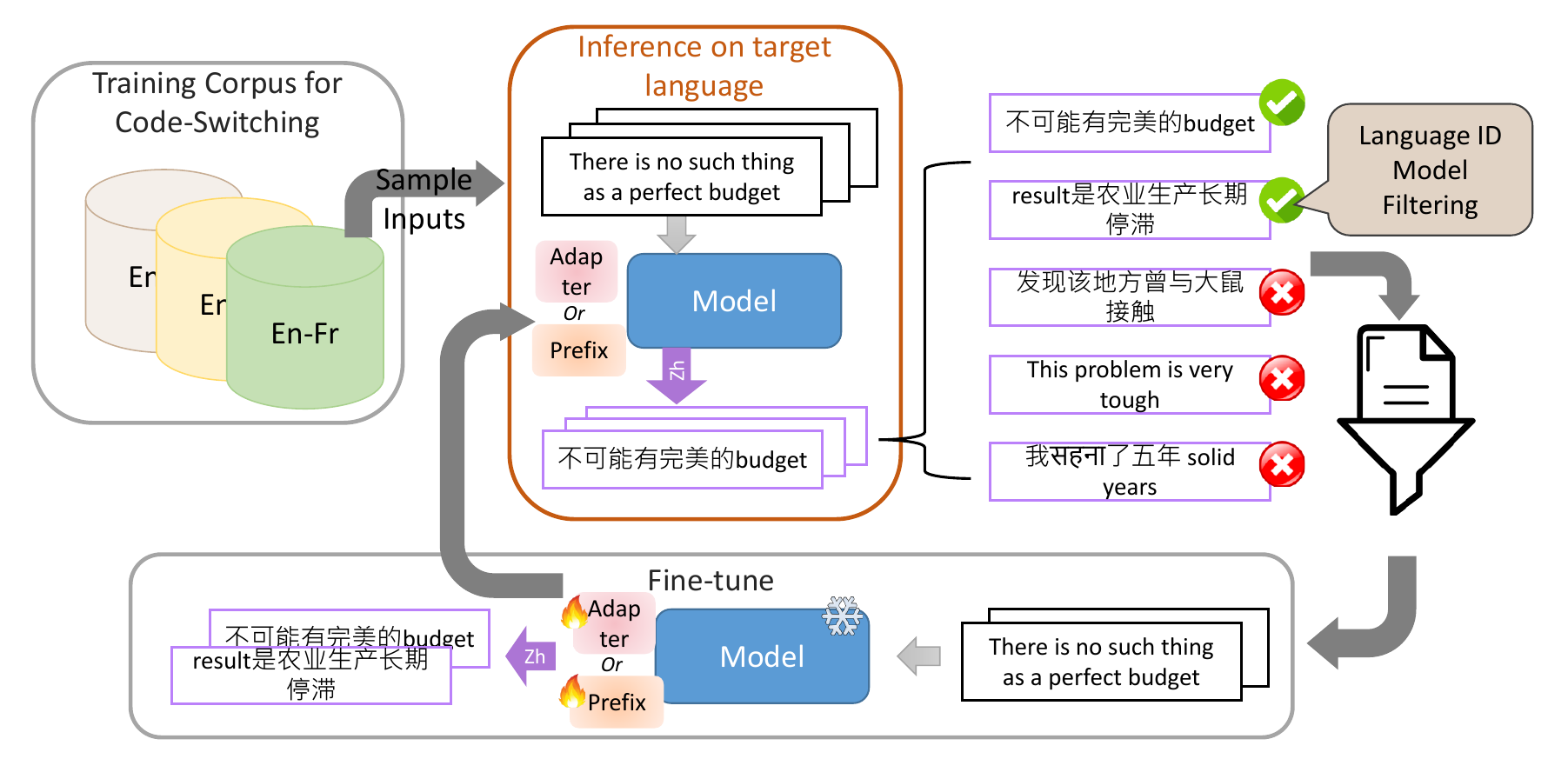}
    \vspace{-0.6em}
    \caption{An illustration of the self-training procedure we designed for \model{}. We use Chinese-English to be the target language pair as an example in this figure.}
    \label{fig:selflearning}
    \vspace{-0.6em}
\end{figure*}

\subsection{PMMTM}
\label{subsec:PMMT}
Multilingual machine translation models~\cite{DBLP:conf/iwslt/HaNW16, DBLP:journals/tacl/JohnsonSLKWCTVW17, DBLP:journals/corr/abs-2205-10835, mbart50} enable simple deployment and parameter-efficient support of machine translation for a large number of language pairs by using a shared representation space. 
To train a PMMTM, we follow the strategy of mBART-50~\cite{mbart50} to notify the model of the source language and the target language to be translated into. Specifically, a language-specific special token is prepended to both the source and target sentences. Hence, during decoding, the first token fed to the decoder is the target language's special token that guides the translation. This is illustrated in \Cref{fig:overview}.

\subsection{The \model{} Model}
\label{subsec:model}
After obtaining a PMMTM, which can comprehend and generate phrases across multiple languages, our next step is to transform a PMMTM into a CS text synthesizer.
A commonly used way is to directly fine-tune the PMMTM on CS training data~\cite{tarunesh2021machine, gupta2020semi}. 
However, models directly fine-tuned on new data could easily overfit to the fine-tuning scenario. Thus it is hard to adapt the ability to perform code-switching to unseen language pairs.
Therefore, instead of directly fine-tuning the whole PMMTM, we propose to use an \textit{additional} code-switching module paired with the PMMTM. The module is specifically learned to \textit{mix} languages for a given translation pair generated by PMMTM.

To implement the design and enable end-to-end training, we employ either an \textit{adapter}~\cite{adapter} or extra \textit{prefixes}~\cite{prefixtuning} as the code-switching module. These approaches are parameter-efficient methods to introduce control into pre-trained models and guide the final generation~\cite{DBLP:conf/iclr/HeZMBN22}:

\paragraph{Adapter} is an additional layer (and parameters) that is introduced inside each Transformer block~\cite{DBLP:conf/nips/VaswaniSPUJGKP17}, and it was shown to be an effective way to conduct transfer learning for NLP tasks~\cite{adapter}. This layer is appended after each feed-forward layer (in a Transformer block). It projects the original feature size to a smaller dimension and then projects them back to the original size, ensuring that the number of parameters stays substantially small.

\paragraph{Prefix} is another parameter-efficient way to conduct transfer learning for NLP tasks~\cite{prefixtuning}. \textit{``Prefix''} are the new key and value matrices used when calculating attention in Transformer. More specifically, trainable prefixes are a set of vectors that will be concatenated with the original key and value matrices when calculating dot-product attention. Hence, in each layer, inputs will be influenced by these additional keys and values after attention is applied.

During fine-tuning using CS training data, we keep the parameters of PMMTM frozen and solely train the adapter or prefixes. This allows the code-switching module to learn how to blend a \textit{translated} distribution with the input sentence. When \model{} is tested and tasked with generating a code-switched sentence in an unseen target language pair, the frozen PMMTM, having been trained to produce translations for this specific pair, can still generate reliable translations. With reliable translations, our code-switching module continues to perform a similar function during training by blending languages. As a result, \model{} exhibits improved generalization capabilities.

\subsection{\model{} with Self-Training}
\label{subsec:self-training}
Although \model{} has the ability to generalize to synthesize CS text to languages that the PMMTM supports, the generation could still be unstable. As we will show in \Cref{sec:human-eval}, \model{} still has around 11\% to 13\% of cases that will generate non-CS sentences when performing zero-shot transfer to unseen language pairs. Hence, we aim to improve this stability issue if more information about the test case is provided.
We assume a common scenario in real practice --- the target language pair $l^m$ and $l^e$ is known, and we can update \model{} for fitting this specific target language pair.

We design a self-training procedure to incorporate off-the-shelf language identification models to help \model{} synthesize target CS sentences more stably. 
The procedure is illustrated in \Cref{fig:selflearning}. To be more specific, we first use the input sentence written in $l^e$ in the CS training data as the input query and ask \model{} to make a prediction on the target language $l^m$, forming potential CS sentences $\mbf x^{m,e}$. Then, we use language identification models to perform sentence filtering based on the following constraints:
\begin{itemize}[topsep=3pt, itemsep=-3pt, leftmargin=13pt]
    \item The synthesized sentence should at least cover one token from $l^m$.
    \item The synthesized sentence should at least cover tokens from $l^e$.
    \item The synthesized sentence cannot cover tokens from other languages except $l^m$ and $l^e$.
\end{itemize}

We use CLD3~\footnote{\url{www.github.com/bsolomon1124/pycld3}} as the language identification model, which extracts character n-grams from the input text and computes an embedding based on the fraction of times each n-gram character appears. 
Notably, CLD3's training does not rely on code-switched text. We leverage CLD3's predicted language distribution for each token to determine if each generated sentence meets the aforementioned constraints. We filter out low-quality instances and collect the remaining sentences as a synthetic code-switching corpus specific to the target domain. This corpus is subsequently used for further fine-tuning of \model{}. The procedure can be executed repeatedly in $R$ rounds, where $R$ is a hyper-parameter. Notice that other advanced filtering can be easily included in our proposed procedure and we leave the exploration as a future work.

Different from the classic self-training algorithm in semi-supervised research~\cite{DBLP:journals/corr/abs-2305-04691}, in our procedure, the initial model is a zero-shot transfer model. Additionally, we apply a filtering process to further improve the quality of the synthetic code-switching corpus.

\subsection{Discussion}
Utilizing pre-trained models that are initially trained on machine translation data as a foundation for constructing code-switched (CS) text synthesizers has gained significant attention recently due to the resemblance between machine translation and CS text synthesis tasks~\cite{tarunesh2021machine, gupta2020semi}. However, our work differs from theirs in that we train a \textit{single} model capable of consuming all the machine translation data, thereby supporting translation across multiple language pairs. In contrast, prior works rely on selecting data based on the target language pair ($l^m$ and $l^e$) as a priori.

Our approach enables a unified model that possesses the ability to generate phrases in multiple languages, thereby facilitating CS text synthesis across various language pairs. Conversely, constraining the training of the PMMTM to a limited number of languages, such as a few specific pairs, would result in \model{} losing its ability to generalize to a broader range of CS language pairs.

\begin{table*}[t!]
\centering
\resizebox{1.0\textwidth}{!}{
\setlength{\tabcolsep}{3.5pt}
\begin{tabular}{llccc|cc|cc|cc}
    \toprule
    \multicolumn{2}{c}{\multirow{2}{*}{\textbf{Model}}} & \multirow{2}{*}{\textbf{Type}} & \multicolumn{2}{c}{\textbf{Bn-En}} & \multicolumn{2}{c}{\textbf{De-En}} & \multicolumn{2}{c}{\textbf{Es-En}} & \multicolumn{2}{c}{\textbf{Hi-En}}\\
    \cmidrule{4-11}
    & & & B & M &B & M &B & M &B & M\\
    \midrule
    \cellcolor{blue!10}&\citet{gupta2020semi}* & Sup.
    & 21.49 & 27.32 & 24.15 & 30.47 & 22.47 & 29.45 & 21.55 & 28.37 \\
    \cellcolor{blue!10}&Fine-tuned PMMTM on all language pairs (mBART50-MMT) & Sup.
    & 12.49 & 38.67 & 32.24 & 59.75 & 37.82 & 62.54 & 27.93 & 54.81 \\
    \cellcolor{blue!10}\multirow{-3}{*}{\rotatebox[origin=c]{90}{\textit{UB.}}}&Fine-tuned PMMTM on all language pairs (augment-MMT) & Sup.
    & 13.08 & 38.69 & 32.65 & 59.96 & 38.59 & 63.36 & 28.88 & 55.10 \\
    &Copy Input & Unsup.
    & 2.66 & 19.28 & 3.29 & 22.76 & 3.28 & 22.31 & 5.22 & 24.20 \\
    &Machine Translation & Unsup.
    & 4.78 & 16.82 & 6.30 & 30.28 & 9.63 & 32.97 & 9.87 & 24.26 \\
    &Translate, Align, then Swap & Unsup.
    & 1.91 & 16.06 & 5.53 & 27.30 & 7.80 & 30.11 & 6.61 & 24.90 \\
    &Fine-tuned PMMTM on available language pairs & Zst.
    & 3.05 & 18.57 & 9.09 & 32.34 & 8.77 & 30.41 & 3.93 & 22.22 \\
    \midrule
    \cellcolor{blue!10}&\model{} (mBART50-MMT + adapter) & Zst.
    & 2.31 & 22.07 & 18.63 & 48.28 & 23.04 & 49.75 & 4.09 & 22.02 \\
    \cellcolor{blue!10}&\model{} (mBART50-MMT + prefix) & Zst.
    & 5.21 & 26.83 & 20.49 & 48.49 & 23.47 & 50.52 & 7.51 & 29.82 \\
    \cellcolor{blue!10}&\model{} (augment-MMT + adapter) & Zst.
    & 2.16 & 18.60 & 14.58 & 40.75 & 16.62 & 42.31 & 8.61 & 30.39 \\
    \cellcolor{blue!10}\multirow{-4}{*}{\rotatebox[origin=c]{90}{\textit{Proposed.}}}&\model{} (augment-MMT + prefix) & Zst.
    & \textbf{9.65} & \textbf{32.63} & \textbf{21.88} & \textbf{50.33} & \textbf{24.85} & \textbf{51.88} & \textbf{12.16} & \textbf{36.94} \\
    \bottomrule
\end{tabular}}
\caption{Automatic evaluation results for \model{}. We evaluate the result in BLEU (B) and METEOR (M). We classify the models into three types --- unsupervised baselines (Unsup.), supervised baselines (Sup.), and zero-shot transfer baselines (Zst.). The training of supervised baselines contains CS data in target language pairs and hence it can be viewed as an upper bound (UB.) for \model{}. Numbers in bold are the best performance among all zero-shot transfer models and unsupervised models. *We report the numbers from the original paper.}
\label{table:auto-main}
\end{table*}
\section{Automatic Evaluation}
\label{sec:auto-eval}
\subsection{Experimental Settings}
\paragraph{Dataset and Evaluation Metrics.}
We use the data provided by \citet{gupta2020semi}, which covers eight language pairs, including Bengali-English (Bn-En), German-English (De-En), Spanish (Es-En), French-English (Fr-En), Hindi-English (Hi-En), Malayalam-English (Ml-En), Tamil-English (Ta-En), and Telugu-English (Te-En). Note that in this dataset, the input language sentence is always English. Hence, the target code-switched (CS) language pair is $X$-English, where $X$ is the different languages that the dataset covers. In the original paper, they used English-$X$ to call the language pair in their dataset, but we changed the naming to present the dominant language first.
The dataset statistics are listed in \Cref{app:dataset}.

In our setting, we conduct leave-one-out experiments, i.e., seven CS language pairs are selected as the CS training data, and the remaining is the test language pair. We select Bn-En, De-En, Es-En, and Hi-En as the four test scenarios based on the language resource levels defined in \citet{mbart50}, such that our selection covers high-resource (German, Spanish), medium-resource (Hindi), and low-resource (Bengali) languages.
We evaluate the synthesized text using BLEU~\cite{DBLP:conf/acl/PapineniRWZ02} and METEOR~\cite{DBLP:conf/acl/BanerjeeL05} scores following \citet{gupta2020semi}.

\begin{table*}[t!]
\centering
\resizebox{1.0\textwidth}{!}{
\setlength{\tabcolsep}{3.5pt}
\begin{tabular}{lcc|cc|cc|cc}
    \toprule
    \multirow{2}{*}{\textbf{Model}} & \multicolumn{2}{c}{\textbf{Bn-En}} & \multicolumn{2}{c}{\textbf{De-En}} & \multicolumn{2}{c}{\textbf{Es-En}} & \multicolumn{2}{c}{\textbf{Hi-En}}\\
    \cmidrule{2-9}
    & B & M &B & M &B & M &B & M\\
    \midrule
    \model{} (mBART50-MMT + prefix)
    & 5.21 & 26.83 & 20.49 & 48.49 & 23.47 & 50.52 & 7.51 & 29.82 \\
    \model{} (mBART50-MMT + prefix) + self-training($R=1$)
    & 5.84 & 28.31 & 20.55 & 48.83 & 24.00 & 51.12 & 8.22 & 31.55 \\
    \model{} (mBART50-MMT + prefix) + self-training($R=2$)
    & \textbf{6.66} & 29.00 & 20.97 & 49.12 & 24.12 & 51.47 & 9.27 & 33.16 \\
    \model{} (mBART50-MMT + prefix)  + self-training($R=5$)
    & 6.26 & \textbf{29.75} & \textbf{21.49} & \textbf{49.71} & \textbf{24.58} & \textbf{51.53} & \textbf{10.31} &\textbf{ 35.84} \\
    \midrule
    \model{} (augment-MMT + prefix)
    & 9.65 & 32.63 & 21.88 & 50.33 & 24.85 & 51.88 & 12.16 & 36.94 \\
    \model{} (augment-MMT + prefix) + self-training($R=1$)
    & 9.80 & 33.63 & 21.78 & 49.73 & 25.96 & 52.68 & 12.99 & 38.59 \\
    \model{} (augment-MMT + prefix) + self-training($R=2$)
    & 10.19 & 34.70 & 22.36 & 50.59 & 26.22 & 52.88 & \textbf{13.70 }& \textbf{40.09} \\
    \model{} (augment-MMT + prefix) + self-training($R=5$)
    & \textbf{10.32} & \textbf{35.46} & \textbf{22.45} & \textbf{50.63} & \textbf{26.31} & \textbf{53.13} & 13.63 & 40.05 \\
    \bottomrule
\end{tabular}}
\caption{Automatic evaluation results for \model{} paired with our self-training procedure. We evaluate the result in BLEU (B) and METEOR (M). Numbers in bold are the best performance among models using the same architecture. We can observe the gradual improvement when more rounds of self-training are applied to \model{}.}
\label{table:self}
\end{table*}

\paragraph{Implementation Details.}
We use two different PMMTM for \model{}. The first one directly adapts the pre-trained mBART50-many-to-many-MMT model (\textbf{mBART50-MMT}) from \cite{mbart50}, which is a machine translation model trained on 50 language pairs using the ML50 benchmark. The other one is to further fine-tune mBART50-MMT on the machine translation data collected by \citet{gupta2020semi} to make an ``augmented mBART50-MMT'' (\textbf{augment-MMT}). 
The second setting is considered since machine translation data in the ML50 benchmark are limited for Indic languages. Hence, we further fine-tune mBART50-MMT on the machine translation data provided in \cite{gupta2020semi} for three epochs. Notice that the machine translation data in \cite{gupta2020semi} only covers eight language pairs, making augment-MMT a more restricted machine translation model in terms of supported languages. 

All \model{} (mBART50-MMT/augment-MMT paired with adapter/prefix) are implemented using the Huggingface package \cite{huggingface} as the backbone. To implement the adapter and prefix, we leverage AdatperHub~\cite{pfeiffer2020AdapterHub}. We use their default setting to set prefix length as 30 and use all prefixes in the self-attention block in the Transformer encoder, and cross-attention block as well as the self-attention block in the Transformer decoder.
We train \model{} with a machine equipped with 4 NVIDIA Tesla V100 GPUs. We train \model{} using 1 GPU at a time with around 30 hrs of training.

We consider AdamW optimizer~\cite{DBLP:conf/iclr/LoshchilovH19} with learning rate set to $10^{-5}$ and the weight decay set to $10^{-5}$. We set the batch size to 12 and the number of training epochs to 15.
For \model{} with self-training, we experiment with $R \in \{1, 2, 5\}$ rounds with heuristics.
Hyper-parameter determination, except for $R$, is based on the available CS data in the development set without considering the leave-out language pair.
Due to the computational resource restriction, our experiment results from a single seed.
We note the gradual performance improvement as $R$ increased in \Cref{subsec:auto-eval-self-training}. However, determining the optimal stopping point for $R$ presented a challenge since no development data exist under the zero-shot scenario. As a result, we decide not to increase $R$ further in our experiments.

\paragraph{Compared baselines.}
Three types of baselines are considered:
\begin{itemize}[topsep=3pt, itemsep=-3pt, leftmargin=13pt]
    \item Unsupervised baselines --- (1) \textbf{Copy Input}: directly copy the input sentence as the prediction, (2) \textbf{Machine Translation}: augment-MMT's machine translation results, (3) \textbf{Translate, Align, then Swap}: we use advanced unsupervised word-alignment tool~\cite{DBLP:conf/eacl/DouN21} to extract potential word alignment between the input sentence and the Machine Translation's prediction. Then, we generate the final output by having a probability $p$ to swap words in Machine Translation's prediction with the aligned input word, where $p=0.35$ is based on the statistics from the training data.
    \item Supervised baselines --- (1) \textbf{\citet{gupta2020semi}}: a sequence-to-sequence model that leverages XLM~\cite{DBLP:conf/nips/ConneauL19} features and utilizes the transfer learning signal from machine translation to warm-up the model, (2) \textbf{Fine-tuned PMMTM on all language pairs}: we fine-tune mBART50-MMT on CS data in \textit{all} eight language pairs. 
    \item Zero-shot transfer baselines --- (1) \textbf{Fine-tuned PMMTM on available language pairs}: fine-tune whole mBART50-MMT on \textit{available} CS training data only (excluding test language pair).

\end{itemize}

Note that the training of supervised baselines contains CS data in target language pairs; hence, it can be viewed as an upper bound for \model{}. Zero-shot transfer baselines are trained only using CS data from other language pairs but not the target language pair. Unsupervised baseline training does not use any CS training data.

\subsection{Main Results} 
\Cref{table:auto-main} shows the results. From the table, we can observe that the unsupervised baselines generate very unreliable CS sentences in general. Additionally, naively fine-tuning the whole PMMTM could perform even worse than the unsupervised methods. \model{} improves unsupervised baselines and zero-shot transfer baselines by at least 55\% relative scores across the board, and every variation of \model{} could outperform these baselines.
By comparing different variations of \model{}, we can observe that \model{} with prefixes is more robust than using an adapter, especially in the cases where the PMMTM model has worse performance (Bengali \& Hindi due to limited training machine translation data used in mBART50-MMT).
Furthermore, by comparing \model{} equipped with augment-MMT and \model{} equipped with mBART50-MMT, we highlight the PMMTM's impact on our model.

\begin{table*}[t!]
\centering
\small
\resizebox{1.0\textwidth}{!}{
\setlength{\tabcolsep}{3.5pt}
\begin{tabular}{llccccc|cccc}
    \toprule
    \multicolumn{2}{c}{\multirow{2}{*}{\textbf{Model}}} & \multirow{2}{*}{\textbf{Type}} & \multicolumn{4}{c|}{\textbf{Hi-En}} & \multicolumn{4}{c}{\textbf{Zh-En}} \\
    \cmidrule{4-11}
    &&& CS Rate. & F & S & Geo. Mean & CS Rate.& F & S & Geo. Mean\\
    \midrule
    &Translate, Align, then Swap
    & Unsup. & 98.6\% & 2.69 & 2.83 & 2.75 & 91.3\% & 3.19 & 3.62 & 3.33 \\
        
    &Fine-tuned PMMTM on available language pairs
    &Zst. & 4.0\% & 1.00 & 1.00 & 1.00 & 2.0\% & 1.0 & 1.0 & 1.0 \\
    
    \cellcolor{blue!10} &\model{} (prefix)
    &Zst. & 87.3\% & 3.06 & 3.10 & 3.08 & 89.3\%& 3.67 & 3.84 & 3.73 \\
    
    \cellcolor{blue!10}\multirow{-2}{*}{\rotatebox[origin=c]{90}{\textit{Ours.}}} & \model{} (prefix + self-training)
    &Zst.& 93.3\% & 3.84 & 3.96 & 3.90 & 99.3\%& 3.73 & 4.01 & 3.85 \\
    \midrule
    \cellcolor{blue!10} &Fine-tuned PMMTM on all language pairs
    &Sup. & 98.0\% & 4.09 & 4.21 & 4.15 & $--$ & $--$ & $--$ & $--$ \\
    
    \cellcolor{blue!10}\multirow{-2}{*}{\rotatebox[origin=c]{90}{\textit{UB.}}}&Ground truth
    &$--$ & 96.0\% & 4.40 & 4.39 & 4.84 & 94.0\% & 4.18 & 4.42 & 4.28 \\
    \bottomrule
\end{tabular}}
\caption{Human evaluation results for \model{} in Hindi-English (Hi-En) and Chinese-English (Zh-En). Code-switching correctness rate (\textit{CS Rate.}) measures the percentage of the prediction is a correct CS. \textit{F} is the abbreviation of the Fluency score. \textit{S} is the abbreviation of the Semantic Correctness score.
Geometric Mean (\textit{Geo. Mean}) is the average of each sample's geometric mean between its code-switching correctness score, fluency and semantic scores. Results for the supervised baseline and the ground truth are presented as an upper bound (UB.) for reference.}
\label{table:human}
\end{table*}

\subsection{Results Given Known Target Language} 
\label{subsec:auto-eval-self-training}
When the target language pair is known, we can then apply our self-training procedure to \model{}. We experiment on \model{} using prefixes and present results in \Cref{table:self}. From the table, we can observe the consistent improvement when adopting self-training to \model{}, and the improvement is especially significant for Hindi-English. Additionally, by conducting self-training with more rounds, we can observe the gradual improvements in both of the cases for \model{} with mBART50-MMT and augment-MMT.

\section{Human Evaluation}
\label{sec:human-eval}
To further verify the quality of our method, we conduct the human evaluation for 
Hindi-English and Chinese-English code-switched (CS) text using sentences in English as the source language.

\subsection{Evaluator Selection}
Considering the expertise of the annotation task requires people familiar with both English and Chinese (or English and Hindi), we have a high-standard selection process to recruit 3 professionals for the human evaluation.
For Hindi-English annotation, We engaged the services of a team of expert professionals who were contracted to provide labels for various Hindi and English-related tasks. They're all native Hindi speakers and highly skilled in speaking Hindi-English code-switching.
Conversely, our Chinese-English annotators are native Chinese NLP researchers with over three years of experience, residing in the US for at least four years, and proficient in Chinese, English, and Chinese-English code-switching. We offer a competitive hourly payment that meets regional legal standards, though it's difficult to determine the average payment for them on this single task.

\subsection{Experimental Settings}
\paragraph{Dataset.}
To avoid the evaluation being biased in the domain we trained on, we collect testing English instances by selecting sentences from the following CS dataset. We sample 50 sentences for each language pair.
\begin{itemize}[topsep=3pt, itemsep=-3pt, leftmargin=13pt]
    \item Hindi-English: We use the data released from \citet{tarunesh2021machine}, who collected the dataset via crowd-sourcing in India. Every data point in this dataset is a pair of an English sentence and its corresponding Hindi-English CS sentence. 
    \item Chinese-English: We use the transcript data of the SEAME dataset~\cite{lyu2010seame}, which is a Chinese-English CS speech recognition dataset. To get the English counterpart of the Chinese-English sentences, we ask experts to translate the CS sentence back to their English version.
\end{itemize}

\begin{figure*}[t!]
    \centering
    \includegraphics[trim=0cm 0cm 0cm 0cm, clip, width=0.95\textwidth]{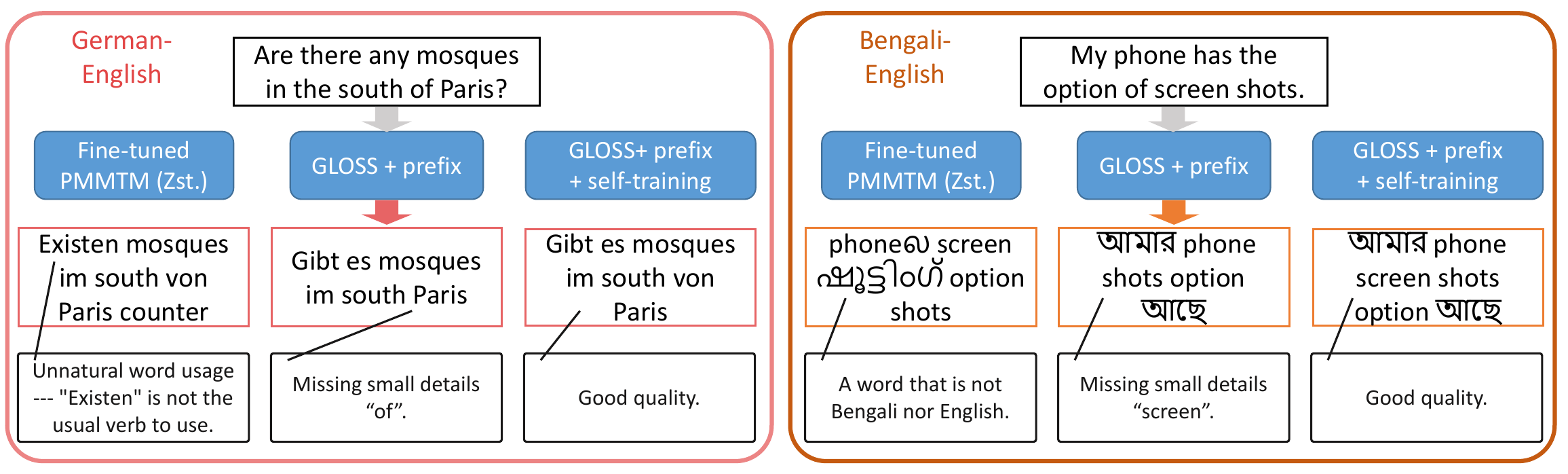}
    \caption{Real examples generated by the models for German-English and Bengali-English cases. Fine-tuned PMMTM (Zst.) refers to the \textit{Fine-tuned PMMTM on available language pairs} method. The explanation for each prediction is presented in the bottom boxes in the Figure.}
    \label{fig:example}
\end{figure*}

\paragraph{Compared models.}
We compare six methods: (1) \textbf{Translate, Align, then Swap}, which serves as a representative of unsupervised methods,  
(2) \textbf{Fine-tuned PMMTM on available language pairs}, which serves as a baseline for zero-shot transfer, 
(3) \textbf{\model{} + prefix}, we use augment-MMT as the backbone for Hindi-English, while using mBART50-MMT as the base model for Chinese-English, 
(4) \textbf{\model{} + prefix + self-training}, we apply self-training ($R=5$) to \model{} + prefix, 
(5) \textbf{Fine-tuned PMMTM on all language pairs}, which serves a strong supervised baseline. Notice that since the training dataset in \citet{gupta2020semi} does not contain the Chinese-English pair. Hence, when evaluating on Chinese-English, this baseline is not applicable,
(6) \textbf{Ground truth}, the original CS sentences we sampled from the dataset.

\paragraph{Evaluation Procedure}
We ask each expert annotator to evaluate all the output of 50 testing instances from all models (i.e., 300 sentences for Hindi-English and 250 for Chinese-English).
Our questionnaire covers the following three questions when using Hindi-English as an example.
\begin{itemize}[topsep=3pt, itemsep=-3pt, leftmargin=13pt]
    \item Code-switching Correctness: We measure whether the present sentence is correct CS (binary score). Specifically, we define a sentence as a correct CS sentence if it satisfies the constraints: (a) It's not fully Hindi or English, (b) It should be mainly in Hindi, and (c) There's no other language except English and Hindi.
    \item Fluency: Measuring the fluency of the prediction presented to humans with scores from 1 to 5, with 5 as the best.
    \item Semantic Correctness: Measuring whether the predicted sentence correctly reveals the meaning of the corresponding input sentence with scores from 1 to 5, with 5 as a fully correct translation.
\end{itemize}

\subsection{Results}
 \Cref{table:human} presents the results. First, we can observe that the code-switching correctness rate is extremely low for the zero-shot baseline --- Fine-tuned PMMTM on available language pairs. Second, although the unsupervised baseline -- Translate, Align, then Swap gets a high code-switching success rate, the low fluency reveals that deciding a suitable position to switch languages is a task beyond random. Third, we can observe that self-training can successfully improve the code-switching quality across all metrics in both languages, indicating the method's effectiveness.

\subsection{Output Examples}
Lastly, we present real examples generated by our models in \Cref{fig:example}. For these examples, we can see that directly fine-tuning the whole PMMTM on CS training data will generate unnatural or even predictions containing tokens in other languages. In contrast, \model{} can generate more stable results, and our self-training algorithm can even help \model{} to generate high-quality CS sentences.

\section{Related Work}
Early approaches~\cite{DBLP:conf/acl/ChoudhuryDBSPB18, DBLP:journals/corr/BhatCB16, DBLP:conf/aclnut/PratapaC21, DBLP:conf/emnlp/LiF14} on code-switched (CS) text synthesis were built based on various linguistic theories, such functional head constraints~\cite{belazi1994code}, Matrix-Language theory~\cite{ myers1997duelling, DBLP:conf/coling/Joshi82}, and Equivalence-Constraint theory~\cite{poplack1980sometimes, sankoff1998formal}. To turn linguistic theories into computational models, \citet{DBLP:journals/corr/BhatCB16, DBLP:conf/aclnut/PratapaC21} leverage trained constituency parser to extract parses of translation pairs and create CS sentences by mixing translation pairs following the syntactic constraints derived from the theories. However, constraints cannot be postulated as a universal rule for all CS scenarios, especially for languages that are syntactically divergent~\cite{berk1986linguistic}, such as English and Chinese, since they have word alignments with an inverted order~\cite{winata2019code}. 
Owing to the limitation, more and more recent works start to build CS text synthesizers in a data-driven way. \citet{DBLP:conf/emnlp/GargPJ18} train a sequence generative adversarial model on real CS text to generate Chinese-English CS sentences. \citet{chang2018code} build a CS text synthesizer using the generative adversarial network, while several follow-up works~\cite{samanta2019deep, winata2019code, DBLP:conf/emnlp/GonenG19} using different generative model techniques are also presented. 
More studies have been introduced to improve the synthesis quality such that we cannot exhaust them in this short summary. We refer readers to the recent survey~\cite{DecadesSurvey2022Winata, DBLP:journals/corr/abs-1904-00784}.

Although many of these efforts had some success, the above-mentioned methods can only generate CS text in the same language pair sets used in training. Given the difficulties of acquiring CS data, this requirement hinders the scalability of these models to support more language pairs. Hence, in this paper, we take a step forward to explore the possibility of zero-shot transfer generalization in CS text synthesis and present \model{} that can generate reasonable outputs.
\section{Conclusion}
\label{sec:conclusion}
In this paper, we develop a novel generalized code-switched text synthesizer, which can even generate code-switched sentences where the corresponding code-switched training data is unavailable. We introduce \model{} that is built on top of a pre-trained multilingual machine translation model and augmented with an adapter or prefixes. The modularized design of learning specific parameters for mixing languages from a translated distribution helps the overall system generalization, hence, fulfilling our goal. 
Extensive experiments verify our methods' effectiveness qualitatively and quantitatively. In the future, we plan to investigate how our synthesizer performs on downstream tasks such as conversational understanding under a code-switched scenario. 
\section*{Limitation}
Our paper presents a pilot exploration of investigating a new setting in code-switched text synthesis --- we allow the target language pair selection not limited to those for which we already have training data. Although we have shown the strength of \model{} qualitatively and quantitatively, our experimental setting is still confined due to the dataset restriction --- all the input text is in English. It would be an even harder challenge if the source languages are more diverse and we leave such exploration for future work.

Additionally, due to the computational restriction, in \model{}, we only explore mBART50-MMT and an augment-MMT as our PMMTM. From the experimental results, we do observe the benefit of having a more stable PMMTM in \model{}. We anticipate the models' performance can be further improved by leveraging more stronger PMMTM, and the exploration is left for the future.

\section*{Broader Impacts}
Our proposed models are based on a model that is pre-trained on a large scale of multilingual machine translation data. It is known that the machine translation model could capture the bias reflecting the training data~\cite{DBLP:conf/acl/WangRC22}. Therefore, our models can potentially generate code-switched text containing offensive or biased content. We suggest that for deploying our model in any real-world applications, careful examination of the potential bias is an essential step.

\section*{Acknowledgements}
The authors would like to thank Chris Hench, Chenyang Tao, Mingyu Derek Ma, Che-Ping Tsai, and Tanmay Parekh for their feedback and help regarding human evaluation. We also thank anonymous reviewers for their helpful feedback on the paper. 

\bibliography{anthology,custom}
\bibliographystyle{acl_natbib}

\clearpage
\appendix

\section{Dataset Details}
\label{app:dataset}
\Cref{tab:data} presents the dataset statistics for our automatic evaluation. The dataset is created by \citet{gupta2020semi} and under a Creative Commons Attribution-NoDerivatives 4.0 International License.\footnote{To download the data, please refer to \url{https://docs.google.com/forms/d/e/1FAIpQLSfR8st2eNu6oq5i499bglxrbJ2BYCQKfpHyaIYq6oS-KrDibA/viewform}.}

\begin{table}[h!]
\centering
\small
\setlength{\tabcolsep}{4pt}
\begin{tabular}{c|ccc}
    \toprule
    Language Pairs & Train & Dev. & Test \\
    \midrule
    Es-En & 196,725 & 2,000 & 2,000 \\
    De-En & 188,131 & 2,000 & 2,000 \\
    Fr-En & 193,922 & 2,000 & 2,000 \\
    Hi-En & 248,330 & 2,000 & 2,000 \\
    Bn-En & 163,893 & 2,000 & 2,000 \\
    Ml-En & 178,453 & 2,000 & 2,000 \\
    Ta-En & 11,380 & 2,000 & 2,000 \\
    Te-En & 9,105 & 2,000 & 2,000 \\
    \bottomrule
\end{tabular}
\caption{Dataset statistics of the dataset provided by \citet{gupta2020semi}.}
\label{tab:data}
\end{table}

\section{Inter Annotator Agreement}
\label{app:human_eval}

We measure the mutual agreement rate among our human annotators by calculating the average absolute differences between the scores they give for the same instance. For example, if the semantic correctness score is given with score $(2, 2, 3)$. Then, the average absolute difference is $0.66$. We then take a micro average across all our human-annotated instances. We get a score of $0.50$ and $0.52$ for the fluency and semantic correctness score for Chinese-English, respectively. As for Hindi-English, we get a score of $0.59$ and $0.55$ for the fluency and semantic correctness score. This indicates our experts agree with each other with only a little disagreement.


\end{document}